\def\year{2021}\relax
\documentclass[letterpaper]{article} 
\usepackage{aaai21}  
\usepackage{times}  
\usepackage{helvet} 
\usepackage{courier}  
\usepackage[hyphens]{url}  
\usepackage{comment}
\usepackage{color}
\usepackage{amsmath}
\usepackage{amsfonts}
\usepackage[switch]{lineno}  %
\usepackage{graphicx} 
\usepackage{natbib}  
\usepackage{caption} 
\usepackage{appendix}

\title{FaceController: Controllable Attribute Editing for Face in the Wild}
\author{
Zhiliang Xu \textsuperscript{\rm 1}\thanks{Authors contribute equally. This work was done when Zhiliang Xu was an intern at Baidu.}, 
Xiyu Yu \textsuperscript{\rm 2}\footnotemark[\value{footnote}], 
Zhibin Hong \textsuperscript{\rm 2}, 
Zhen Zhu \textsuperscript{\rm 1}, 
Junyu Han \textsuperscript{\rm 2}, \\
Jingtuo Liu \textsuperscript{\rm 2}, 
Errui Ding \textsuperscript{\rm 2}, 
Xiang Bai \textsuperscript{\rm 1}\thanks{Corresponding author}
}
\affiliations{

    \textsuperscript{\rm 1} Huazhong University of Science and Technology,
    \textsuperscript{\rm 2}  Baidu Inc. \\
    \{zhiliangxu1, zzhu, xbai\}@hust.edu.cn, \{yuxiyu, hongzhibin, hanjunyu, liujingtuo,  dingerrui\}@baidu.com

}

\begin{document}

\maketitle

\begin{abstract}

Face attribute editing aims to generate faces with one or multiple desired face attributes manipulated while other details are preserved. Unlike prior works such as GAN inversion, which has an expensive reverse mapping process, we propose a simple feed-forward network to generate high-fidelity manipulated faces. By simply employing some existing and easy-obtainable prior information, our method can control, transfer, and edit diverse attributes of faces in the wild. The proposed method can consequently be applied to various applications such as face swapping, face relighting, and makeup transfer. In our method, we decouple identity, expression, pose, and illumination using 3D priors; separate texture and colors by using region-wise style codes. All the information is embedded into adversarial learning by our identity-style normalization module. Disentanglement losses are proposed to enhance the generator to extract information independently from each attribute. Comprehensive quantitative and qualitative evaluations have been conducted. In a single framework, our method achieves the best or competitive scores on a variety of face applications.

\end{abstract}

\section{Introduction}
Manipulating and editing real faces in the wild has countless applications in the areas of visual effects and e-commerce. Face editing tasks often benefit from the highly disentangled latent representation of face attributes such that one can precisely manipulate corresponding attributes. For example, face swapping, a special case of face editing, aims to preserve the identity of a source face while retaining the pose and expression of a target face. Therefore, achieving fortified disentanglement of face attribute latent codes is the core of many works~\cite{zhang2019multi,shen2020interpreting,ha2019marionette,zeng2020realistic}.

However, deriving such disentangled latent representation from a real face is still challenging. Directly mapping a face into the latent space by an encoder~\cite{chen2018isolating,klys2018learning} can be efficient. But it is not ensured to extract latent codes that express semantically-meaningful factors of variations~\cite{nie2020semi}. Thus, another common practice is to obtain the reverse mapping in latent space for a real face, known as \emph{GAN inversion}. For example, \cite{abdal2019image2stylegan,zhu2020domain} opt to embed a real face into the latent space of StyleGAN~\cite{StyleGAN} by optimizing a randomly initialized code through back-propagation. In practice, approaches of this kind are often time-consuming and cumbersome. 

Inspired by recent face synthesis methods such as~\cite{gecer2018semi,deng2020disentangled}, we propose a simple feed-forward face generation network that is fed with some existing and easy-obtainable prior information. As such, this method can avoid the costly learning process of disentangled representation. In this paper, we name our method as FaceController, which enables the control of face attributes. We embed 3DMM coefficients into an adversarial learning framework. However, we observe there exists a domain gap between real faces and the counterpart ones rendered from 3DMM. For example, the makeup of a woman cannot be perfectly expressed by 3DMM. To combat this, we further employ the embedding of identity and region-wise style codes as side information to complement the information gap between 3DMM coefficients and real faces, making our method stand out from previous works.

In the proposed FaceController, a new identity-style normalization module is proposed to fuse all the prior information. The identity-style normalization module is an extension of SPADE~\cite{SPADE} in which we use the embedding of identity and region-wise style codes to modulate the feature maps originated from 3DMM coefficients. Furthermore, to make the generator generalize to different editing, we devise unsupervised face generation losses on identity, landmarks, and textures to enforce the generator to control each attribute independently.

FaceController provides more comprehensive functions of real face editing by precisely controlling the latent codes of a wide range of face attributes. Unlikely previous methods, which mainly address the disentanglement of attributes such as identity and expression, identity and pose, our method is able to disentangle in total five different attributes, including identity, pose, expression, illumination, and local region styles. As a result, FaceController enables a variety of face applications. For example, it can transfer the lip color or eye shadow makeup by editing the region-wise codes; it can also be applied to adjust the illumination of a given face image. In this paper, we empirically validate that the proposed method achieves better quantitative scores and visual results in most of the above applications.

The main contributions of this paper are:
(1) FaceController generates high-quality face with desired attributes using a simple feed-forward face generation network and improves efficiency by employing easily-obtainable prior information rather than the time-consuming reverse mapping process; (2) FaceController also increases the freedom of disentangled and controllable dimensions of attributes, making it better to control, transfer, and edit face attributes. 

\section{Related Work}

\subsubsection{Disentangled representation learning.}  In face attribute editing, many attempts have been made to learn an independent representation of face attributes.
Early attempts include unsupervised disentangled latent representation learning of real face images. For example,  InfoGAN~\cite{chen2016infogan} and its variants~\cite{lin2019infogan} target to maximize the mutual information between latent representation. HoloGAN~\cite{nguyen2019hologan} learns 3D representation from natural images to disentangle identity and pose.
StyleGAN~\cite{StyleGAN} can synthesize highly photorealistic face images and allow scale-specific modifications to the styles. Variational Autoencoder (VAE) based methods~\cite{shen2017learning,razavi2019generating,chen2018isolating,sun2018mask} often encourage independence of the latent representation by regularizing their
total correlation. These methods achieved a promising performance. However, the disentangled latent codes are not provably identifiable~\cite{locatello2019challenging}, and strong regularization on latent representation sometimes may be harmful to generate high-fidelity face images due to information loss~\cite{he2019attgan}.

Other attempts aim to extract disentangled latent representation using some explicit or implicit supervision information \cite{bouchacourt2017multi}. For example,  ~\cite{deng2020disentangled,tewari2020stylerig} aim to manipulate the StyleGAN latent space by using 3DMM information as supervision during learning. ~\cite{shen2020interpreting} tries to classify the latent space into multiple linear subspaces with semantic meaning. It is also possible to impose explicit supervision on latent codes~\cite{nie2020semi}. Most of these works can manipulate fake faces with high precision. However, in order to edit real faces, an optimization-based embedding process named GAN inversion has to be conducted to infer the best latent code for this face. It is often time-consuming. 

\subsubsection{Facial manipulation using 3D priors.} 
3DMMs~\cite{blanz1999morphable,cao2013facewarehouse} represent the shape and appearance of faces by projecting them into low-dimensional spaces using techniques such as Principal Component Analysis (PCA). 3DMMs provide an explicit way to independently represent information such as identity, expression, and pose. Therefore, it can be naturally applied to face manipulation. For example, Face2Face~\cite{thies2016face2face} models both the driving and source face via a 3DMM; then applies the expression components of driving face to source face. ~\cite{kim2018deep,xu2020deep} try to manipulate expression and pose by modifying 3DMM coefficients. ~\cite{gecer2018semi} proposes a semi-supervised adversarial framework for generating photorealistic faces directly from 3DMM. However, we observe that simply deploying 3D priors often results in a domain gap between real faces and generated faces due to the information loss of 3DMM. To minimize this domain gap, we propose an identity-style normalization module to fuse some side information. Consequently, we can obtain more photorealistic faces.

\section{Controllable Attribute Editing}

\begin{figure*}[t!]
	\centering
	\includegraphics[width=0.99\textwidth]{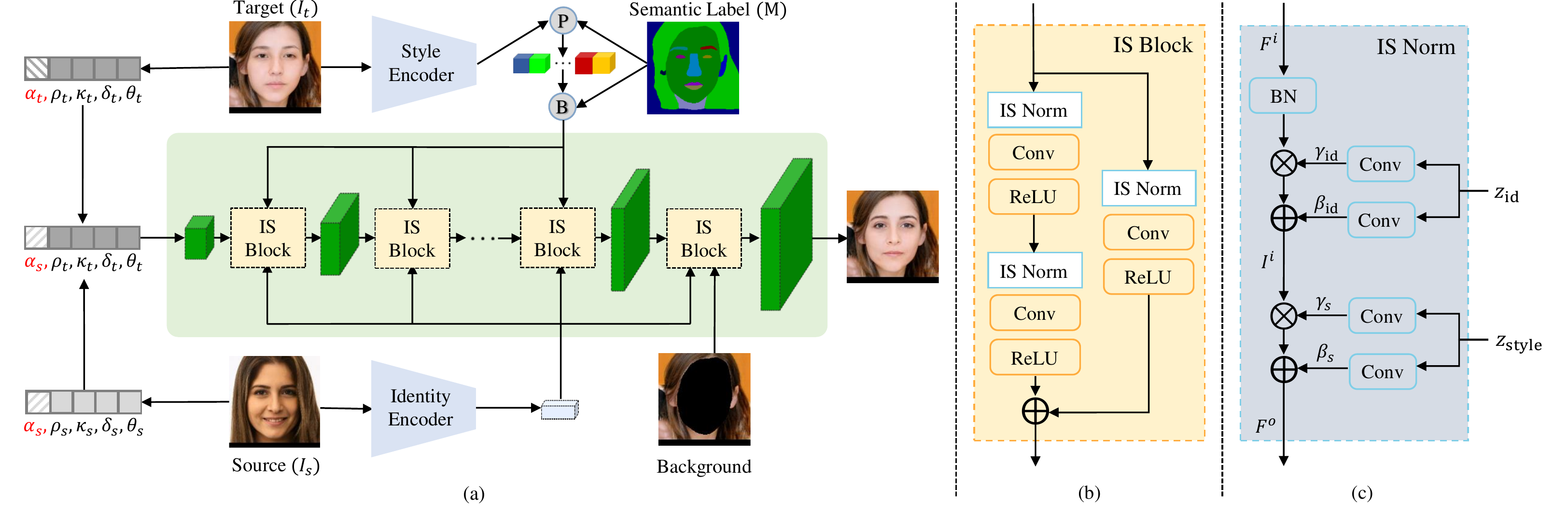}
	\caption{(a) The architecture of FaceController. We take the task of face swapping as an example. The 3DMM coefficients $(\alpha_s, \rho_s, \kappa_s, \delta_s, \theta_s)$ and $(\alpha_t, \rho_t, \kappa_t, \delta_t, \theta_t)$ are extracted from source and target images $I_s$ and $I_t$, respectively. First, we transfer the coefficients of identity from source to target and obtain $(\alpha_s, \rho_t, \kappa_t, \delta_t, \theta_t)$. The transferred 3DMM coefficients with the style and identity codes are fed into the generator to synthesis a swapped face. Here, $P$ denotes region-wise pooling; and $B$ denotes the broadcasting process which is the inverted operation of pooling. (b) The architecture of IS block. IS block contains several Identity-Style Normalization modules that integrate all the information of above latent representation. (c) The architecture of IS normalization module. $z_\mathrm{id}$ and $z_\mathrm{style}$ represent identity and region-wise style respectively.}
	\label{fig:network}
\end{figure*}

In this paper, we aim to efficiently edit faces in the wild with desired attributes. We propose a simple feed-forward framework to generate the edited faces directly from some easily obtainable latent representation of attributes. Specifically, denote $I$ as a face image and $\{a_1, \cdots, a_n\}$ as its latent representation of $n$ different attributes. With replacing $a_i, (i \in {1, \cdots, n})$ by desired latent code $a_i'$, we can generate desired face image $I'$ and achieve attribute editing.

To achieve this, FaceController requires at least three stages: 1) an encoding stage that aims to produce attribute-independent latent codes; 2) an attribute editing stage that modifies the produced latent codes to the desired ones through simple replacement; 3) a decoding stage that receives the edited latent codes as input and produces face images that correspond to the edited latent codes. 

FaceController has specific and useful designs for stronger disentanglement of a wide range of face attributes and interpreting them to high-quality and plausible face images that conform to the edited latent codes. Taking face swapping as an example, the architecture of FaceController generator is shown in Fig..~\ref{fig:network} (a). Each part of FaceController is built with clear and reasonable motivations and narrated as follows.

\subsection{Face Attribute Disentanglement}
Among available solutions to perform face attribute disentanglement, 3DMM~\cite{egger20203d} stands as a near-optimal option because it disentangles a wide range of different attributes including identity, expression, pose, illumination, and texture. Many previous works~\cite{deng2020disentangled} conduct attributes disentanglement based on 3DMM coefficients and show impressive results. However, we notice that the expressive power of its \emph{identity} and \emph{texture} coefficients are not sufficient to reconstruct a perceptually photo-realistic face. This drawback attributes to the limited volume of 3D face data where the original 3DMM~\cite{deng2019accurate} is trained, limiting its generalization ability to real-world faces. Considering this, we believe identity embedding and region-wise texture representation should be included to complement the insufficient 3DMM identity and texture representation. A recent work~\cite{zhu2020sean} particularly centers on region-wise face control while it is not designed for other attributes control. To successfully modulate 3DMM attributes control and region-wise styles control in a unified architecture, we first extract 3DMM coefficients and then build two encoders that encode more representative identity and region-wise styles, respectively. 

\subsubsection{3DMM cofficients extraction.}

Specifically in a 3DMM, a face is often modeled with 3D shape $\mathcal{S}$ and texture $\mathcal{T}$, which is obtained by:
\begin{equation}
\begin{aligned}
    \mathcal{S} &= \mathcal{\overline{S}} + \alpha I_{\mathrm{base}} + \rho E_{\mathrm{base}}, \\
    \mathcal{T} &= \mathcal{\overline{T}} + \delta T_{\mathrm{base}}.
\end{aligned}
\end{equation}
In the above equation, $\mathcal{\overline{S}}$ and $\mathcal{\overline{T}}$ are the average face shape and texture, respectively; $I_{\mathrm{base}}$, $E_{\mathrm{base}}$, and $T_{\mathrm{base}}$ are the PCA bases of identity, expression, and texture, respectively; $\alpha \in \mathbb{R}^{80}$, $\rho \in \mathbb{R}^{64}$, and $\delta \in \mathbb{R}^{80}$ are the corresponding 3DMM coefficients for rendering a 3D face. 
Besides shape and texture, the illumination $\kappa \in \mathbb{R}^{27}$ of the face can be approximated with Spherical Harmonics (SH) and the pose is defined as three rotation angles. For simplicity, we use $\theta$ to denote the parameters of rotation and translation. Luckily, $(\alpha, \rho,  \kappa, \delta, \theta)$ can be obtained by an off-the-shelf 3D face reconstruction network~\cite{deng2019accurate} and we use this network as the 3DMM coefficients extractor.

\subsubsection{Identity encoder.} 
As discussed previously, 3DMM is not able to perfectly depict the identity of a wild face. To complement the identity information, we employ an identity encoder to extract identity information from real images. Here, a state-of-the-art pretrained face recognition model~\cite{deng2019arcface} is used as 
the identity encoder. 
We use the last feature maps before the final fully-connected (FC) layer to obtain a precise and high-level representation of identity. It is worthy to note that the face images fed into the 3DMM coefficients extractor and identity encoder are aligned in a different way. Then, in our identity encoder, we introduce the spatial transformation network~\cite{jaderberg2015spatial} to provide a precisely aligned face for our identity encoder. 

\subsubsection{Region-wise style encoder.} 
To support local face region editing, we aim to obtain region-wise style codes by taking advantage of the semantic segmentation of face images. A segmented region that is assigned with the same class label associates with a style code. We adopt the encoder of SEAN~\cite{zhu2020sean} as our region-wise style encoder for its superior ability to extract region-wise styles. Specifically, a region pooling layer is applied at the end of the encoder for each semantic label to extract a 512 dimensional style code. This region-wise style encoder naturally provides a disentangled representation of style for local face regions.

\subsection{Face Attribute Remapping}
At this point, we have acquired different sorts of latent codes---3DMM coefficients, identity embedding, and region-wise style codes---for detailed and precise attribute disentanglement. The next is how to leverage these codes and decode them back to a plausible generated face image.

Our task is generally related to translate semantic labels to natural face images with specific region-wise styles. SPADE~\cite{SPADE} is especially useful to this task and some recent works SEAN~\cite{zhu2020sean} and SMIS~\cite{zhu2020semantically} have built their methods based on SPADE to support region-wise editing. Based on this observation, we also take the spirit of SPADE to design our decoder. However, our task requires the decoder to additionally consider region-irrelevant attributes such as identity. To support this, we design an \emph{Identity-Style Normalization module} that integrates the identity and region-wise style information to the decoder, shown in Fig.~\ref{fig:network} (c).

Let $F^i \in \mathbb{R}^{B\times C^i \times H^i \times W^i}$ denote the feature maps of the $i$-th layer of the decoder; $B$, $C^i$, $H^i$, and $W^i$ represent the batch size, channel numbers, height, and width, respectively. We first perform the batch normalization on $F^i$ and get $\overline{F^i}$. To integrate identity information, we map the identity embedding to a vector $z_\mathrm{id}$ with the same dimension of channel numbers by using an FC layer. After that, two normalization parameters, $\gamma_\mathrm{id}$ and $\beta_\mathrm{id}$, are learned from this vector to modulate $\overline{F^i}$ like a style code. The modulated output is,

\begin{equation}
    I^i = \gamma_\mathrm{id} \circ \overline{F^i} + \beta_\mathrm{id},
\end{equation}
where $\gamma_\mathrm{id}$ and $\beta_\mathrm{id}$ are with the same size of $\overline{F^i}$; we denote $\circ$ as the element-wise production.

To further integrate region-wise style codes, we first extract a style matrix $S \in \mathbb{R}^{B\times N \times D}$ for region style codes by our region-wise style encoder. We also provide a corresponding segmentation map $M^i \in \mathbb{R}^{B\times N \times H^i \times W^i}$ which is interpolated from the segmentation map of the input face. Here, $N$ denotes the number of semantic classes; $D$ is the dimension of style code for each semantic class. Such segmentation map has $N$ channels as every channel is responsible for a single class. To adaptively adjust a specific local region by its corresponding style code, we broadcast the style code into a feature map according to its segmentation mask. This broadcast process can be achieved by applying a batch-version matrix multiplication between the segmentation map $M^i$ and style matrix $S$, i.e.,
\begin{equation}
    z_\mathrm{style} = S^{\top} \times M^i,
\end{equation}
where $S^{\top}\in \mathbb{R}^{B\times D\times N}$ denotes the transposed matrix of $S$; $z_\mathrm{style}\in \mathbb{R}^{B\times D\times H^i\times W^i}$ are the broadcast style codes. Similar to identity normalization, we learn two parameters $\gamma_{s}$ and $\beta_{s}$ from $z_\mathrm{style}$ to modulate $I^i$, then we have the output of IS normalization module,
\begin{equation}
    F^{o} = \gamma_{s} \circ {I^i} + \beta_{s}.
\end{equation}

\subsection{Training Losses}
During training, we mainly consider two different training processes, \emph{i.e.}, face reconstruction and unsupervised face generation training. During face reconstruction training, the model receives attribute latent codes from the same face image and tries to reconstruct such a face. In this case, $I_s$ is equal to $I_t$ as depicted in Fig.~\ref{fig:network}. Such training ensures that the generated image is reasonable when $I_s = I_t$. However, the model is not ensured to work well when $I_s \neq I_t$. Consequently, we include unsupervised face generation training to guarantee even with unpair input ($I_s \neq I_t$), the model is still able to translate the combined latent codes to a plausible face image. Note that GAN loss is applied to both training processes for better image fidelity. 

\subsubsection{Face Reconstruction Loss.}
Following FaceShifter~\cite{li2019faceshifter}, we utilize $20\%$ training data for face reconstruction. We implement face reconstruction loss using the perceptual loss~\cite{johnson2016perceptual} to supervise the reconstruction. The intermediate feature maps of the generated face $I_g$ and the given face $I_t$ are extracted from a pre-trained VGG network~\cite{simonyan2014very}, and then we conduct pixel-level reconstruction as follows,
\begin{equation}
    \mathcal{L}_\mathrm{per} = \frac{1}{2} \| \mathcal{F}_\mathrm{per}(I_t) - \mathcal{F}_\mathrm{per}(I_g) \|^2,
\end{equation}
where $\mathcal{F}_\mathrm{per}$ denotes the extractor of feature maps. 

\subsubsection{Unsupervised Face Generation Loss.}
The main purpose of our work is to support free and dynamic face attribute control over face images. Basically, when we alter some attributes, we wish to see results with these attributes changed accordingly while other attributes are unchanged. We design \emph{identity loss}, \emph{face landmark loss} and \emph{histogram matching loss} to assist the unsupervised training and further fortify attribute disentanglement.


If we aim to transfer the attributes such as expression and pose from $I_t$ to $I_s$ while keeping the identity, we can encourage the generated face to keep the same identity with $I_s$ by utilizing the identity encoder:
\begin{equation}
    \mathcal{L}_\mathrm{id} =  1-\cos(\mathcal{F}_\mathrm{id}(I_g), \mathcal{F}_\mathrm{id}(I_s)).
\end{equation}
Here, $\mathcal{L}_\mathrm{id}$ refers to the identity loss and $\mathcal{F}_\mathrm{id}$ is the extractor of identity embedding. We use cosine similarity to estimate the similarity between the identity embedding of the generated image and the source image.

\begin{figure}[t!]
	\centering
	\includegraphics[width=0.47\textwidth]{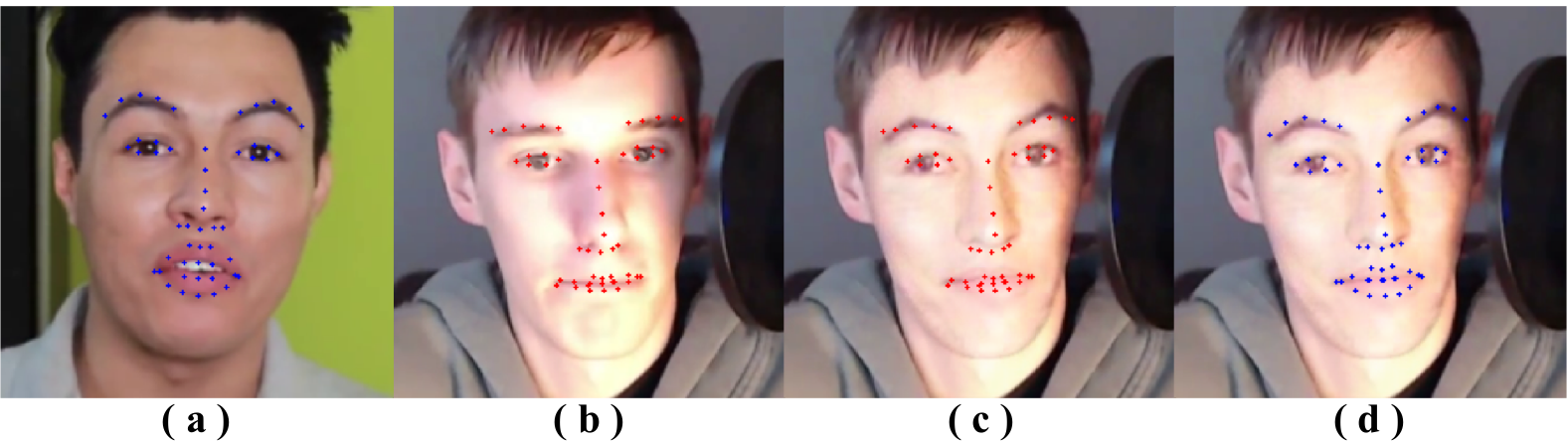}
	\caption{Demonstration of our landmark loss. Our aligned landmarks can better preserve the identity of the source image. (a) Source face with its landmarks. (b) Target face with its landmarks. (c) The swapped face with the landmarks of target face. As we can see, the landmarks and the face image are not well aligned (Zoom view better). (d) The swapped face with our aligned landmarks.  }
	\label{fig:landmark}
\end{figure}

Then, if we want to change attributes like identity and textures from $I_s$ to $I_t$ but retain the expression and pose, we can leverage the landmark loss to ensure expression and pose consistency between $I_t$ and the generated face $I_g$:
\begin{equation}
    \mathcal{L}_\mathrm{lm} = \frac{1}{2}\| \mathcal{F}_\mathrm{lm}(I_g) - \mathcal{F}_\mathrm{lm}(I_t) \|^2,
\end{equation}
where $\mathcal{F}_\mathrm{lm}$ represents the face landmark extractor.

In this paper, the landmark loss is also specifically designed for identity preservation. As we want to keep the generated image $I_g$ having the same expression and pose with target image $I_t$, a common implementation is to constrain the $I_g$ and $I_t$ to have the same landmarks. However, face landmarks contain the shape of the eyes, mouth, eyebrows, and nose. These are also linked with identity information. As shown in Fig.~\ref{fig:landmark} (a) and Fig.~\ref{fig:landmark} (b), different persons have different face features and landmarks. Consequently, as demonstrated in Fig.~\ref{fig:landmark} (c), using landmarks of $I_t$ to constrain the generated image is not a good way as we want to preserve identity information of $I_s$. We need to adjust the landmarks to have the same pose and expression with $I_t$ while keeping the same identity information with $I_s$. To solve this problem, we utilize the swapped 3DMM coefficients as shown in Fig.~\ref{fig:network} to extract 3D landmarks. These landmarks often can precisely keep the same face features with $I_s$, seen in Fig.~\ref{fig:landmark} (d). Then, matching the 3D landmarks of $I_g$ and the aligned landmarks can better preserve the identity information.


Even though identity loss can maintain correct identity, consistency in local textures and colors is still hard to keep. To solve this problem, we further introduce a histogram matching loss~\cite{risser2017stable} to encourage region-wise style consistency with a target face.
Let $I_{re} = \mathrm{HM}(I_t, I_g)$ be a remapped image obtained by the histogram matching between $I_g$ and $I_t$. Then, we have
\begin{equation}
    \mathcal{L}_\mathrm{hm} = \frac{1}{2}\| I_g - I_\mathrm{re} \|^2.
\end{equation}
For this loss to be minimized, the color distribution of $I_g$ is learned to be similar to $I_t$, making local styles of the generated face more similar to the target.

\subsubsection{Full Loss} 
During training, the full loss is defined as a weighted sum of above losses:
\begin{equation}
    \mathcal{L} = \mathcal{L}_\mathrm{adv} + \lambda_\mathrm{id}\mathcal{L}_\mathrm{id} +  \lambda_\mathrm{lm} \mathcal{L}_\mathrm{lm} + \lambda_\mathrm{hm} \mathcal{L}_\mathrm{hm} + \lambda_\mathrm{per}\mathcal{L}_\mathrm{per},
\end{equation}
where $\mathcal{L}_\mathrm{adv}$ denotes GAN loss. We set the $\lambda_\mathrm{id}=10$, $\lambda_\mathrm{lm}=10000$, $\lambda_\mathrm{hm}=100$, and $\lambda_\mathrm{per}=100$, respectively.

\section{Experiments}
\begin{figure}[t!]
	\centering
	\includegraphics[width=0.47\textwidth]{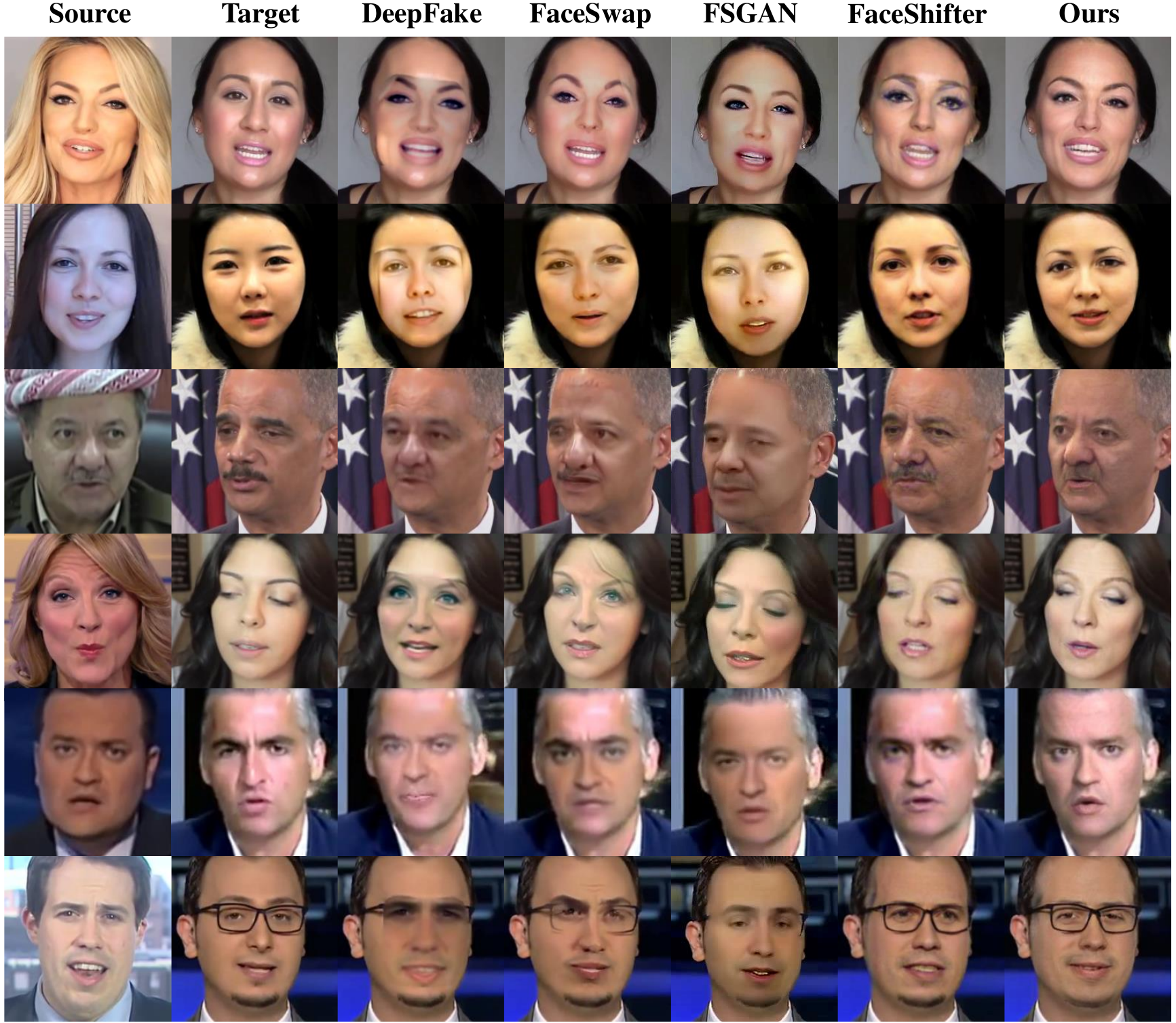}
	\caption{Comparison with DeepFake, FaceSwap, FSGAN, and FaceShifter on FaceForensics++ dataset.}
	\label{fig:faceforensics}
\end{figure}

\subsubsection{Implementation details.} The training images for FaceController are collected from CelebA-HQ~\cite{karras2017progressive}, FFHQ~\cite{StyleGAN}, and VGGFace~\cite{parkhi2015deep} datasets. The face alignment for 3DMM coefficients extraction is processed according to \cite{deng2019accurate}. The size of an aligned face is $224 \times 224$. For region-wise style codes, we employ an off-the-shelf BiSeNet model~\cite{yu2018bisenet} to obtain semantic labels, which provides 19 different region categories. The generator and discriminator are trained around 500K steps, respectively. More details can be found in the Supplementary Materials.


\subsection{Qualitative Analysis}

\begin{figure*}[t!]
	\centering
	\includegraphics[width=1.0\textwidth]{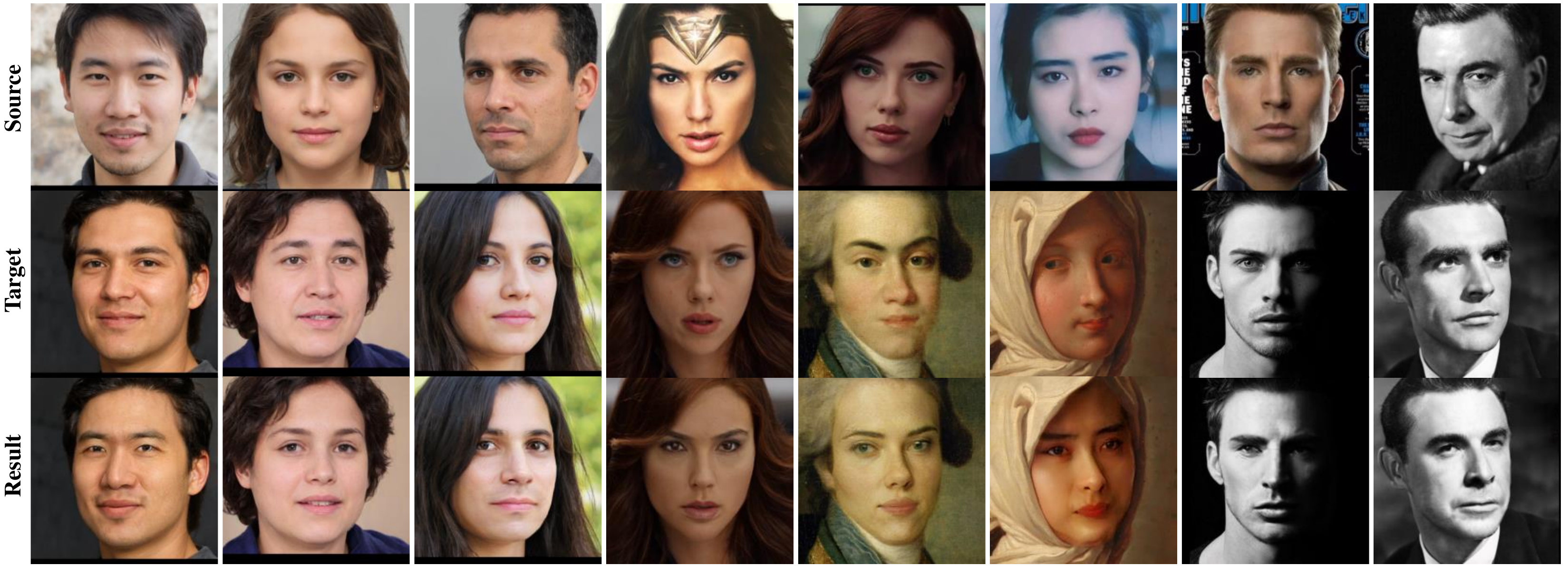}
	\caption{Face swapping on different sources of face images. Here, the first three images are generated by StyleGAN2 and others are collected from the Internet.}	
	\label{fig:faceswap}
\end{figure*}

\begin{figure}[t!]
	\centering
	\includegraphics[width=0.47\textwidth]{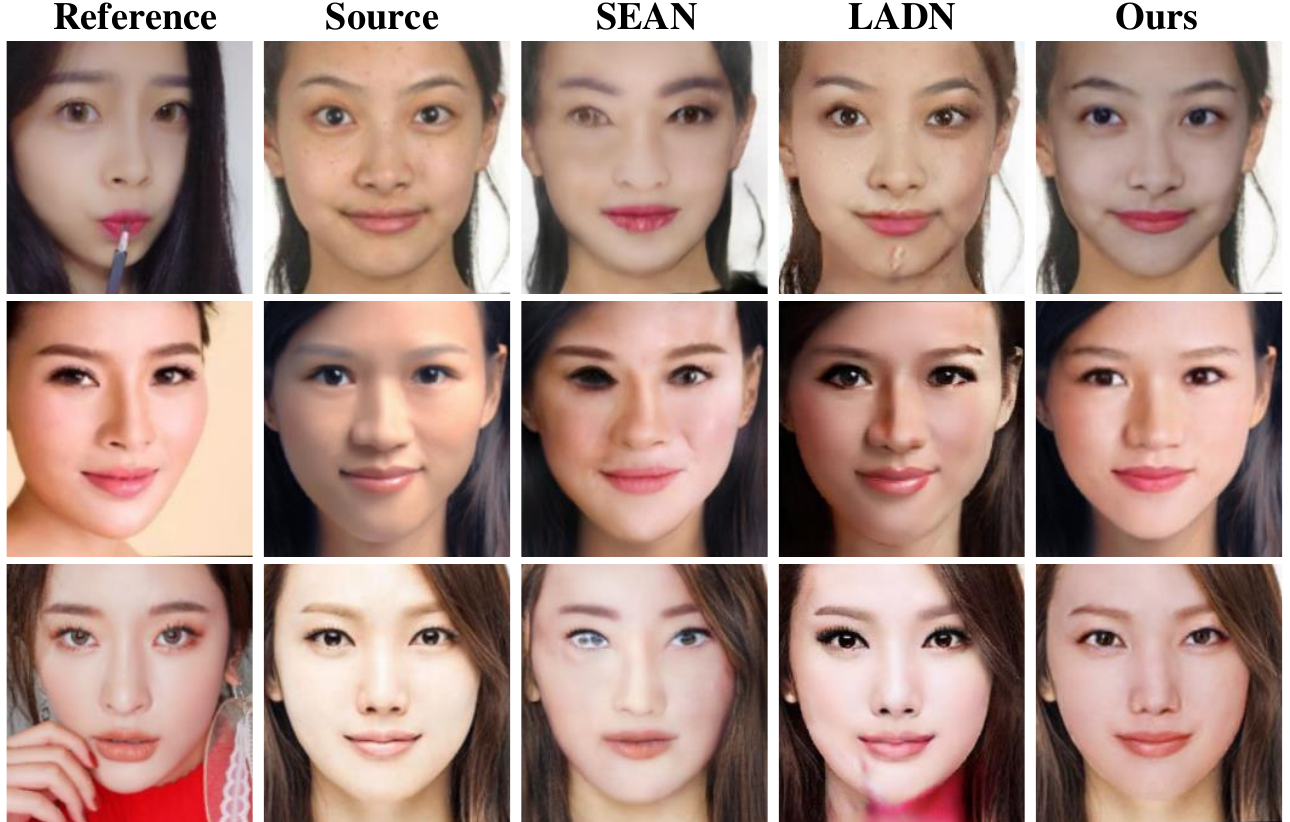}
	\caption{Comparison with other makeup transfer methods.}
	\label{fig:makeup_comparison}
\end{figure}

\subsubsection{Face swapping.} 
We first apply our method to the task of face swapping. Our method is compared with recent state-of-the-art methods including DeepFake~\cite{petrov2020deepfacelab}, FaceSwap\footnote{\url{https://github.com/MarekKowalski/FaceSwap}}, FSGAN~\cite{nirkin2019fsgan}, and FaceShifter~\cite{li2019faceshifter}.All methods are evaluated on the FaceForensics++ dataset~\cite{rossler2019faceforensics}. As shown in Fig.~\ref{fig:faceforensics}, 
even though the rival methods are specifically designed for this problem, our method still obtains better or comparable results. 
Note that some methods, such as DeepFake and FSGAN, include a blending process to fuse the swapped faces with the backgrounds of target faces. Such a blending process is unstable and prone to failure, resulting in unpleasant illumination and artifacts. As contrast, our method and FaceShifter discard the blending process and achieve better results.
Compared to FaceShifter, our method shows better preservation of identity because we have explicitly enhanced the disentanglement of attributes by carefully designing the representation and learning losses.

To demonstrate the power of our method on editing diverse face images, we also evaluate our method on face images downloaded from the Internet and faces generated by StyleGAN2~\cite{karras2020analyzing}. The results are shown in Fig.~\ref{fig:faceswap}. We observe that our method achieves very impressive results even when swapping faces with uncommon illumination or textures.


\subsubsection{Region-wise style editing.} 
\begin{figure}[t!]
	\centering
	\includegraphics[width=0.47\textwidth]{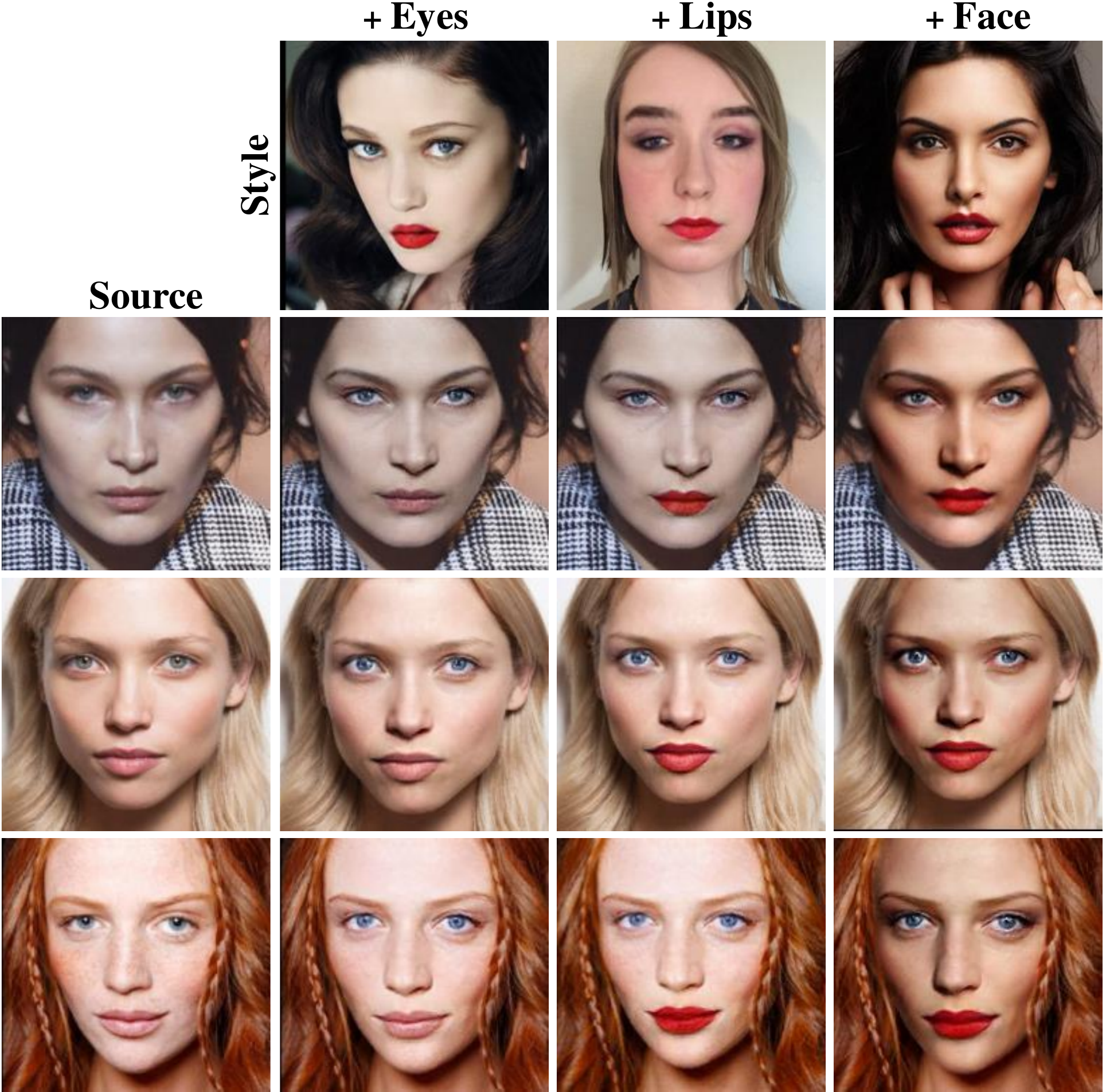}
	\caption{Progressive makeup transfer.}
	\label{fig:makeup_our}
\end{figure}

The disentanglement of region-wise style codes enables our method to achieve high performance in makeup transfer. To validate this, we compare with recent methods such as LADN~\cite{gu2019ladn} and SEAN~\cite{zhu2020sean}. All methods are evaluated on MT-dataset~\cite{dantcheva2012can}. As illustrated in Fig.~\ref{fig:makeup_comparison}, our method appears to preserve more identity as compared to SEAN. Our method also ensures the fine-grained makeup transfer by utilizing region-wise style codes to adaptively adjust different face regions. This cannot be achieved by LADN, which uses a single entangled code to represent the whole face. To further demonstrate our capability of fine-grained makeup transfer, we show
more results in Fig.~\ref{fig:makeup_our}. Here, we progressively transfer the styles of eyes, lips, and faces based on the results of the previous column. This can be simply implemented by editing corresponding region-wise style codes. These results also further verify the strong disentanglement of region-wise style codes.

\subsubsection{Face manipulation and relighting.}
\begin{figure}[t!]
	\centering
	\includegraphics[width=0.47\textwidth]{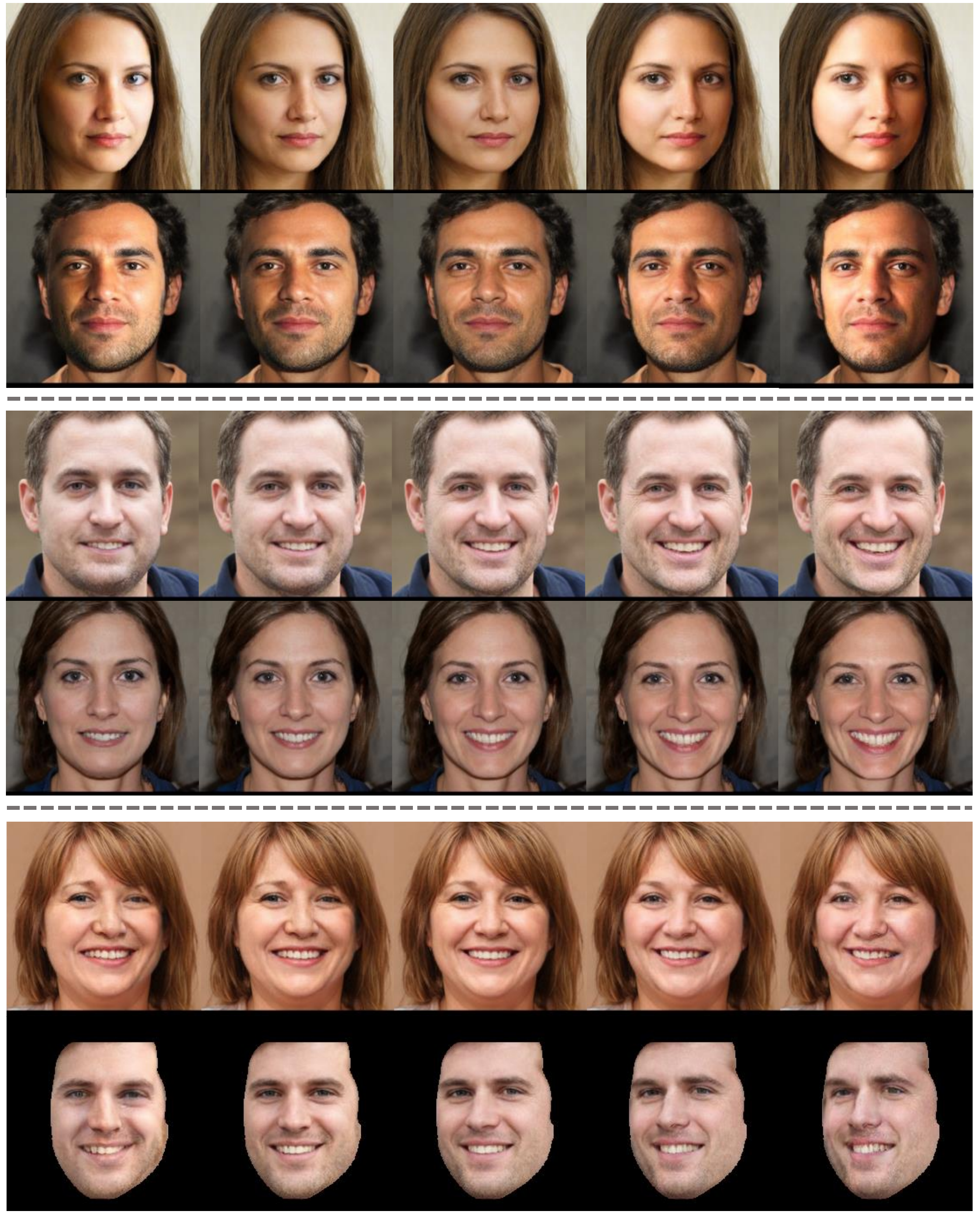}
	\caption{The interpolated results with respect to illumination, expression, and poses from top to bottom.}
	\label{fig:coeff_all}
\end{figure}

Our method can independently adjust the expression, pose, and illumination of faces. We demonstrate this ability by interpolating the above factors independently. The results are presented in Fig~\ref{fig:coeff_all}. Based on the visual inspection of these results, we observe our method enables precise manipulation of the illumination and expression. To obtain a photorealistic background, we directly embed the background information into the generator rather than enforcing the network to learn how to inpaint. Therefore, The generated image may appear black holes along the edge of the face when the pose varies hugely.

\begin{figure}[t!]
	\centering
	\includegraphics[width=0.47\textwidth]{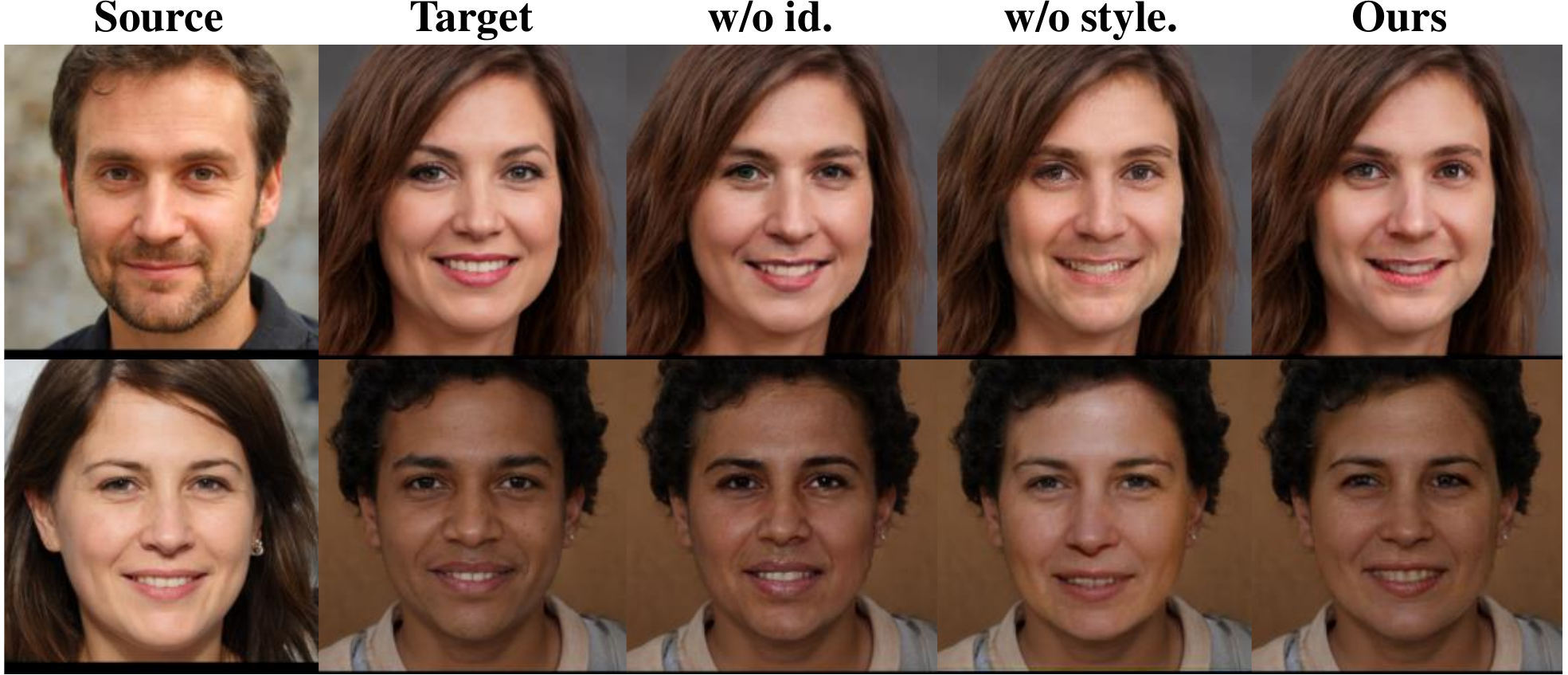}
	\caption{The face swapping results in ablation study. Here, w/o id. refers to our generator without identity encoder; w/o style. refers to our generator without style encoder.}
	\label{fig:ablation}
\end{figure}

\subsection{Quantitative Analysis}
\subsubsection{Face swapping. } We report the quantitative results on FaceForensics++ dataset and compare our method with DeepFake, FaceSwap, FSGAN, and FaceShifter. We follow the metrics in FaceShifter. We uniformly sample 10 frames from each of 1000 videos and get 10K faces in total. Then, we evaluate the accuracy of identity retrieval (abbrev. Retr.) to check whether the identity of source face is preserved after face swapping; the expression (abbrev. Exp.) and pose error to check whether the swapped face keeps the expression and pose of the target face. We also use FID scores~\cite{heusel2017gans} to evaluate the fidelity of swapped faces.

For identity retrieval, we apply CosFace~\cite{wang2018cosface} to extract identity embedding and retrieve the closest face by using cosine similarity. For pose and expression evaluation, we use a pose estimator to estimate head pose and an expression recognition model to extract expression embedding. Then, we report the L2 distance of pose vectors and expression embedding. The pose estimator is the same with that in FaceShifter. Since the expression model used in FaceShifter is not open-source, we employ another model, i.e.~\cite{meng2019frame}, to enable expression error comparison. As shown in Tab.~\ref{tab:faceswap}, our method achieves the best results in identity preservation and fidelity. 


\subsubsection{Region-wise style editing.} We evaluate the performance of makeup transfer by users study. We mainly consider three dimensions: 
(1) the identity preservation; (2) the makeup transferability; (3) the fidelity of faces. 
All methods generate 1K images with the same source and reference faces. Five users with rich knowledge in image generation are asked to rate the best result in each case under each metric. Finally, the average scores are reported in Tab.~\ref{tab:makeup}. Since LADN is specially designed to disentangle the identity and makeup, it enables the excellent preservation of identity and makeup. But compared with LADN, our method can generate higher-quality faces, as shown in Fig.~\ref{fig:makeup_comparison}. SEAN can perform very well on fine-grained makeup transfer. But it cannot precisely preserve the identity and often contains unnatural textures which affects its scores in all aspects.


\begin{table}[t!]
\centering
\begin{tabular}{lcccc}
\hline
Method         & Retr. &  Pose &  Exp. & FID \\ \hline
DeepFake       &     81.96           &    4.14    &      0.47     &   4.29  \\
FaceSwap       &    54.19            &    \textbf{2.51}       &    0.40      &  3.81  \\ 
FSGAN          &     57.34        &     2.84   &        0.38         &   4.35           \\
FaceShifter    &       97.38         &    2.96    &        \textbf{0.36}      &  4.05   \\ \hline
Ours           &      \textbf{98.27}          &      2.65  &        0.39      &  \textbf{3.51}  \\ \hline
\end{tabular}
\caption{The accuracies of identity retrieval, pose and expression errors, and FID scores of different face swapping methods on FaceForensics++ dataset.}
\label{tab:faceswap}
\end{table}

\begin{table}[t!]
\centering
\begin{tabular}{lccc}
\hline
Method     & ID     &   Style     &  Fidelity \\ \hline
SEAN       &   6.2     &      14.1     &   5.8  \\
LADN       &   \textbf{49.1}     &   40.5      &    38.8     \\ \hline
Ours       &    44.7    &      \textbf{45.4}    &    \textbf{55.4}     \\ \hline
								
\end{tabular}
\caption{Users study results w.r.t. identity preservation, makeup transfer, and fidelity.}
\label{tab:makeup}
\end{table}

\begin{table}[t!]
\centering
\begin{tabular}{lcccc}
\hline
Method              &  Retr. &   Pose  & Exp.   &  FID \\ \hline
w/o id.             &       25.97       &     \textbf{2.57}    &  \textbf{0.36}       &    4.12  \\ 
w/o style.          &       98.21       &     2.83     &      0.41        &   4.84    \\ \hline
Ours                & \textbf{98.27}    &     2.65     &      0.39         &    \textbf{3.51}   \\ \hline
								
\end{tabular}
\caption{The accuracies of identity retrieval, pose and expression errors, and FID scores in ablation experiments.}
\label{tab:ablation}
\end{table}

\subsection{Ablation Study}
\begin{figure}[t!]
	\centering
	\includegraphics[width=0.47\textwidth]{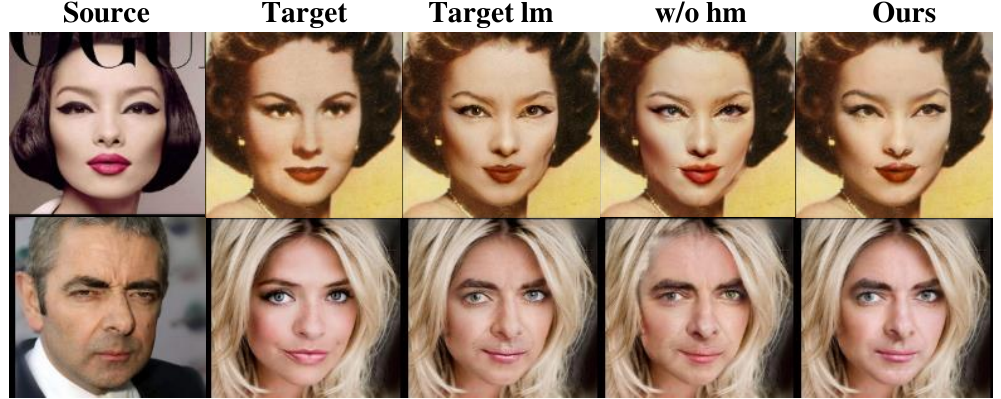}
	\caption{The ablation study about loss functions. Here, Target lm refers we use target image's landmark as ground truth. w/o hm refers to drop out histogram matching loss.}
	\label{fig:ablation_loss}
\end{figure}

\subsubsection{Face representation.} To verify the necessity of the identity encoder and style encoder, we conduct ablation experiments: the proposed generator with (1) no style encoder and (2) no identity encoder. Taking face swapping as an example, the results are demonstrated in Fig.~\ref{fig:ablation}. If we discard the identity embedding, the generated images have identity gaps with source images, as shown in the third column of Fig.~\ref{fig:ablation}. Further, when we discard the style encoder, the generated styles aren't consistent with target image in regions like lips, as shown in the fourth column of Fig.~\ref{fig:ablation}. This indicates that 3DMM coefficients don't provide detailed texture for generating high-fidelity faces. The quantitative results are also consistent with 
the above conclusion. When style information isn't given, the identity is nearly not affected due to the disentangled representation. When identity information isn't provided, the accuracy of identity retrieval severely drops. Note that identity encoder and style encoder are irrelevant to pose and expression change, the pose and expression errors of these two models are very close to our full model.
\subsubsection{Loss functions.}
To verify the benefit of our landmark and histogram matching loss, we make comparisons with (1) using the target image's landmark as ground truth. (2) dropping out the histogram matching loss. The results can be seen in Fig.~\ref{fig:ablation_loss}. If we use the target image's landmark as ground truth, the eyes and the eyebrows are not consistent with the source image. The generated image also has artifacts near the landmark's area. If we drop out the histogram matching loss, the generated image's color is not consistent with the target image. As a comparison, our method can generate better results.   

\section{Limitations and Conclusion}
Currently, our approach has some limitations that can be addressed in follow-up work. As we mentioned above, when changing pose hugely, the generated image may appear black holes. The gaze direction may also be not consistent with the target image. This can be optimized by adding landmarks of the eyeball to constrain the gaze direction. 

In conclusion, we propose a feed-forward network to generate photorealistic faces with desired attributes. This network designs a disentangled face representation and devises unsupervised losses to ensure the control of various face attributes. A variety of face applications can be implemented by this method.  

\section{Acknowledgments}
This work was supported by the National Program for Support of Top-notch Young Professionals and the Program for HUST Academic Frontier Youth Team 2017QYTD08, which are given to Dr. Xiang Bai.
{\fontsize{9.0pt}{10.0pt} \selectfont
\bibliography{main}

\begin{thebibliography}{51}
\providecommand{\natexlab}[1]{#1}
\providecommand{\url}[1]{\texttt{#1}}
\providecommand{\urlprefix}{URL }
\expandafter\ifx\csname urlstyle\endcsname\relax
  \providecommand{\doi}[1]{doi:\discretionary{}{}{}#1}\else
  \providecommand{\doi}{doi:\discretionary{}{}{}\begingroup
  \urlstyle{rm}\Url}\fi

\bibitem[{Abdal, Qin, and Wonka(2019)}]{abdal2019image2stylegan}
Abdal, R.; Qin, Y.; and Wonka, P. 2019.
\newblock Image2{S}tyle{GAN}: How to embed images into the {S}tyle{GAN} latent
  space?
\newblock In \emph{Proc. CVPR}, 4432--4441.

\bibitem[{Blanz, Vetter et~al.(1999)}]{blanz1999morphable}
Blanz, V.; Vetter, T.; et~al. 1999.
\newblock A morphable model for the synthesis of 3{D} faces.
\newblock In \emph{Proc. SIGGRAPH}, volume~99, 187--194.

\bibitem[{Bouchacourt, Tomioka, and Nowozin(2018)}]{bouchacourt2017multi}
Bouchacourt, D.; Tomioka, R.; and Nowozin, S. 2018.
\newblock Multi-level variational autoencoder: Learning disentangled
  representations from grouped observations.
\newblock In \emph{Proc. AAAI}.

\bibitem[{Cao et~al.(2013)Cao, Weng, Zhou, Tong, and
  Zhou}]{cao2013facewarehouse}
Cao, C.; Weng, Y.; Zhou, S.; Tong, Y.; and Zhou, K. 2013.
\newblock Face{W}are{H}ouse: A 3{D} facial expression database for visual
  computing.
\newblock \emph{TVCG} 20(3): 413--425.

\bibitem[{Chen et~al.(2018)Chen, Li, Grosse, and Duvenaud}]{chen2018isolating}
Chen, R.~T.; Li, X.; Grosse, R.~B.; and Duvenaud, D.~K. 2018.
\newblock Isolating sources of disentanglement in variational autoencoders.
\newblock In \emph{Proc. NeurIPS}, 2610--2620.

\bibitem[{Chen et~al.(2016)Chen, Duan, Houthooft, Schulman, Sutskever, and
  Abbeel}]{chen2016infogan}
Chen, X.; Duan, Y.; Houthooft, R.; Schulman, J.; Sutskever, I.; and Abbeel, P.
  2016.
\newblock Info{GAN}: Interpretable representation learning by information
  maximizing generative adversarial nets.
\newblock In \emph{Proc. NeurIPS}, 2172--2180.

\bibitem[{Dantcheva, Chen, and Ross(2012)}]{dantcheva2012can}
Dantcheva, A.; Chen, C.; and Ross, A. 2012.
\newblock Can facial cosmetics affect the matching accuracy of face recognition
  systems?
\newblock In \emph{Proc. BTAS}, 391--398. IEEE.

\bibitem[{Deng et~al.(2019{\natexlab{a}})Deng, Guo, Xue, and
  Zafeiriou}]{deng2019arcface}
Deng, J.; Guo, J.; Xue, N.; and Zafeiriou, S. 2019{\natexlab{a}}.
\newblock Arcface: Additive angular margin loss for deep face recognition.
\newblock In \emph{Proc. CVPR}, 4690--4699.

\bibitem[{Deng et~al.(2020)Deng, Yang, Chen, Wen, and
  Tong}]{deng2020disentangled}
Deng, Y.; Yang, J.; Chen, D.; Wen, F.; and Tong, X. 2020.
\newblock Disentangled and Controllable Face Image Generation via 3{D}
  Imitative-Contrastive Learning.
\newblock In \emph{Proc. CVPR}, 5154--5163.

\bibitem[{Deng et~al.(2019{\natexlab{b}})Deng, Yang, Xu, Chen, Jia, and
  Tong}]{deng2019accurate}
Deng, Y.; Yang, J.; Xu, S.; Chen, D.; Jia, Y.; and Tong, X. 2019{\natexlab{b}}.
\newblock Accurate 3{D} face reconstruction with weakly-supervised learning:
  From single image to image set.
\newblock In \emph{Proc. CVPR Workshops}, 0--0.

\bibitem[{Egger et~al.(2020)Egger, Smith, Tewari, Wuhrer, Zollhoefer, Beeler,
  Bernard, Bolkart, Kortylewski, Romdhani et~al.}]{egger20203d}
Egger, B.; Smith, W.~A.; Tewari, A.; Wuhrer, S.; Zollhoefer, M.; Beeler, T.;
  Bernard, F.; Bolkart, T.; Kortylewski, A.; Romdhani, S.; et~al. 2020.
\newblock 3{D} Morphable Face Models—Past, Present, and Future.
\newblock \emph{TOG} 39(5): 1--38.

\bibitem[{Gecer et~al.(2018)Gecer, Bhattarai, Kittler, and Kim}]{gecer2018semi}
Gecer, B.; Bhattarai, B.; Kittler, J.; and Kim, T.-K. 2018.
\newblock Semi-supervised adversarial learning to generate photorealistic face
  images of new identities from 3{D} morphable model.
\newblock In \emph{Proc. ECCV}, 217--234.

\bibitem[{Gu et~al.(2019)Gu, Wang, Chiu, Tai, and Tang}]{gu2019ladn}
Gu, Q.; Wang, G.; Chiu, M.~T.; Tai, Y.-W.; and Tang, C.-K. 2019.
\newblock L{ADN}: Local adversarial disentangling network for facial makeup and
  de-makeup.
\newblock In \emph{Proc. ICCV}, 10481--10490.

\bibitem[{Ha et~al.(2019)Ha, Kersner, Kim, Seo, and Kim}]{ha2019marionette}
Ha, S.; Kersner, M.; Kim, B.; Seo, S.; and Kim, D. 2019.
\newblock Mario{NET}te: Few-shot face reenactment preserving identity of unseen
  targets.
\newblock In \emph{Proc. AAAI}.

\bibitem[{He et~al.(2019)He, Zuo, Kan, Shan, and Chen}]{he2019attgan}
He, Z.; Zuo, W.; Kan, M.; Shan, S.; and Chen, X. 2019.
\newblock Att{GAN}: Facial attribute editing by only changing what you want.
\newblock \emph{TIP} 28(11): 5464--5478.

\bibitem[{Heusel et~al.(2017)Heusel, Ramsauer, Unterthiner, Nessler, and
  Hochreiter}]{heusel2017gans}
Heusel, M.; Ramsauer, H.; Unterthiner, T.; Nessler, B.; and Hochreiter, S.
  2017.
\newblock G{AN}s trained by a two time-scale update rule converge to a local
  nash equilibrium.
\newblock In \emph{Proc. NeurIPS}, 6626--6637.

\bibitem[{Jaderberg et~al.(2015)Jaderberg, Simonyan, Zisserman
  et~al.}]{jaderberg2015spatial}
Jaderberg, M.; Simonyan, K.; Zisserman, A.; et~al. 2015.
\newblock Spatial transformer networks.
\newblock In \emph{Proc. NeurIPS}, 2017--2025.

\bibitem[{Johnson, Alahi, and Fei-Fei(2016)}]{johnson2016perceptual}
Johnson, J.; Alahi, A.; and Fei-Fei, L. 2016.
\newblock Perceptual losses for real-time style transfer and super-resolution.
\newblock In \emph{Proc. ECCV}, 694--711. Springer.

\bibitem[{Karras et~al.(2017)Karras, Aila, Laine, and
  Lehtinen}]{karras2017progressive}
Karras, T.; Aila, T.; Laine, S.; and Lehtinen, J. 2017.
\newblock Progressive growing of gans for improved quality, stability, and
  variation.
\newblock \emph{arXiv preprint arXiv:1710.10196} .

\bibitem[{Karras, Laine, and Aila(2019)}]{StyleGAN}
Karras, T.; Laine, S.; and Aila, T. 2019.
\newblock A Style-Based Generator Architecture for Generative Adversarial
  Networks.
\newblock In \emph{Proc. CVPR}, 4401--4410.

\bibitem[{Karras et~al.(2020)Karras, Laine, Aittala, Hellsten, Lehtinen, and
  Aila}]{karras2020analyzing}
Karras, T.; Laine, S.; Aittala, M.; Hellsten, J.; Lehtinen, J.; and Aila, T.
  2020.
\newblock Analyzing and improving the image quality of {S}tyle{GAN}.
\newblock In \emph{Proc. CVPR}, 8110--8119.

\bibitem[{Kim et~al.(2018)Kim, Garrido, Tewari, Xu, Thies, Niessner, P{\'e}rez,
  Richardt, Zollh{\"o}fer, and Theobalt}]{kim2018deep}
Kim, H.; Garrido, P.; Tewari, A.; Xu, W.; Thies, J.; Niessner, M.; P{\'e}rez,
  P.; Richardt, C.; Zollh{\"o}fer, M.; and Theobalt, C. 2018.
\newblock Deep video portraits.
\newblock \emph{TOG} 37(4): 1--14.

\bibitem[{Klys, Snell, and Zemel(2018)}]{klys2018learning}
Klys, J.; Snell, J.; and Zemel, R. 2018.
\newblock Learning latent subspaces in variational autoencoders.
\newblock In \emph{Proc. NeurIPS}, 6444--6454.

\bibitem[{Li et~al.(2020)Li, Bao, Yang, Chen, and Wen}]{li2019faceshifter}
Li, L.; Bao, J.; Yang, H.; Chen, D.; and Wen, F. 2020.
\newblock Face{S}hifter: Towards high fidelity and occlusion aware face
  swapping.
\newblock In \emph{Proc. CVPR}.

\bibitem[{Lin et~al.(2019)Lin, Thekumparampil, Fanti, and Oh}]{lin2019infogan}
Lin, Z.; Thekumparampil, K.~K.; Fanti, G.; and Oh, S. 2019.
\newblock Info{GAN}-{CR}: Disentangling Generative Adversarial Networks with
  Contrastive Regularizers.
\newblock \emph{arXiv preprint arXiv:1906.06034} .

\bibitem[{Locatello et~al.(2019)Locatello, Bauer, Lucic, Raetsch, Gelly,
  Sch{\"o}lkopf, and Bachem}]{locatello2019challenging}
Locatello, F.; Bauer, S.; Lucic, M.; Raetsch, G.; Gelly, S.; Sch{\"o}lkopf, B.;
  and Bachem, O. 2019.
\newblock Challenging common assumptions in the unsupervised learning of
  disentangled representations.
\newblock In \emph{Proc. ICML}, 4114--4124.

\bibitem[{Meng et~al.(2019)Meng, Peng, Wang, and Qiao}]{meng2019frame}
Meng, D.; Peng, X.; Wang, K.; and Qiao, Y. 2019.
\newblock Frame attention networks for facial expression recognition in videos.
\newblock In \emph{Proc. ICIP}, 3866--3870. IEEE.

\bibitem[{Nguyen-Phuoc et~al.(2019)Nguyen-Phuoc, Li, Theis, Richardt, and
  Yang}]{nguyen2019hologan}
Nguyen-Phuoc, T.; Li, C.; Theis, L.; Richardt, C.; and Yang, Y.-L. 2019.
\newblock Holo{GAN}: Unsupervised learning of 3{D} representations from natural
  images.
\newblock In \emph{Proc. ICCV}, 7588--7597.

\bibitem[{Nie et~al.(2020)Nie, Karras, Garg, Debhath, Patney, Patel, and
  Anandkumar}]{nie2020semi}
Nie, W.; Karras, T.; Garg, A.; Debhath, S.; Patney, A.; Patel, A.~B.; and
  Anandkumar, A. 2020.
\newblock Semi-Supervised {S}tyle{GAN} for Disentanglement Learning.
\newblock \emph{arXiv} arXiv--2003.

\bibitem[{Nirkin, Keller, and Hassner(2019)}]{nirkin2019fsgan}
Nirkin, Y.; Keller, Y.; and Hassner, T. 2019.
\newblock F{SGAN}: Subject Agnostic Face Swapping and Reenactment.
\newblock In \emph{Proc. ICCV}, 7184--7193.

\bibitem[{Park et~al.(2019)Park, Liu, Wang, and Zhu}]{SPADE}
Park, T.; Liu, M.; Wang, T.; and Zhu, J. 2019.
\newblock Semantic Image Synthesis With Spatially-Adaptive Normalization.
\newblock In \emph{Proc. CVPR}, 2337--2346.

\bibitem[{Parkhi, Vedaldi, and Zisserman(2015)}]{parkhi2015deep}
Parkhi, O.~M.; Vedaldi, A.; and Zisserman, A. 2015.
\newblock Deep face recognition.
\newblock In \emph{Proc. BMVC}.

\bibitem[{Petrov et~al.(2020)Petrov, Gao, Chervoniy, Liu, Marangonda, Um{\'e},
  Jiang, RP, Zhang, Wu et~al.}]{petrov2020deepfacelab}
Petrov, I.; Gao, D.; Chervoniy, N.; Liu, K.; Marangonda, S.; Um{\'e}, C.;
  Jiang, J.; RP, L.; Zhang, S.; Wu, P.; et~al. 2020.
\newblock Deep{F}ace{L}ab: {A} simple, flexible and extensible face swapping
  framework.
\newblock \emph{arXiv preprint arXiv:2005.05535} .

\bibitem[{Razavi, van~den Oord, and Vinyals(2019)}]{razavi2019generating}
Razavi, A.; van~den Oord, A.; and Vinyals, O. 2019.
\newblock Generating diverse high-fidelity images with {VQ}-{VAE}-2.
\newblock In \emph{Proc. NeurIPS}, 14866--14876.

\bibitem[{Risser, Wilmot, and Barnes(2017)}]{risser2017stable}
Risser, E.; Wilmot, P.; and Barnes, C. 2017.
\newblock Stable and controllable neural texture synthesis and style transfer
  using histogram losses.
\newblock \emph{arXiv preprint arXiv:1701.08893} .

\bibitem[{Rossler et~al.(2019)Rossler, Cozzolino, Verdoliva, Riess, Thies, and
  Nie{\ss}ner}]{rossler2019faceforensics}
Rossler, A.; Cozzolino, D.; Verdoliva, L.; Riess, C.; Thies, J.; and
  Nie{\ss}ner, M. 2019.
\newblock Face{F}orensics++: Learning to detect manipulated facial images.
\newblock In \emph{Proc. ICCV}, 1--11.

\bibitem[{Shen and Liu(2017)}]{shen2017learning}
Shen, W.; and Liu, R. 2017.
\newblock Learning residual images for face attribute manipulation.
\newblock In \emph{Proc. CVPR}, 4030--4038.

\bibitem[{Shen et~al.(2020)Shen, Gu, Tang, and Zhou}]{shen2020interpreting}
Shen, Y.; Gu, J.; Tang, X.; and Zhou, B. 2020.
\newblock Interpreting the latent space of {GAN}s for semantic face editing.
\newblock In \emph{Proc. CVPR}, 9243--9252.

\bibitem[{Simonyan and Zisserman(2014)}]{simonyan2014very}
Simonyan, K.; and Zisserman, A. 2014.
\newblock Very deep convolutional networks for large-scale image recognition.
\newblock \emph{arXiv preprint arXiv:1409.1556} .

\bibitem[{Sun et~al.(2018)Sun, Huang, Shi, and Ma}]{sun2018mask}
Sun, R.; Huang, C.; Shi, J.; and Ma, L. 2018.
\newblock Mask-aware photorealistic face attribute manipulation.
\newblock \emph{arXiv preprint arXiv:1804.08882} .

\bibitem[{Tewari et~al.(2020)Tewari, Elgharib, Bharaj, Bernard, Seidel,
  P{\'e}rez, Zollhofer, and Theobalt}]{tewari2020stylerig}
Tewari, A.; Elgharib, M.; Bharaj, G.; Bernard, F.; Seidel, H.-P.; P{\'e}rez,
  P.; Zollhofer, M.; and Theobalt, C. 2020.
\newblock Style{R}ig: Rigging {S}tyle{GAN} for 3{D} Control over Portrait
  Images.
\newblock In \emph{Proc. CVPR}, 6142--6151.

\bibitem[{Thies et~al.(2016)Thies, Zollhofer, Stamminger, Theobalt, and
  Nie{\ss}ner}]{thies2016face2face}
Thies, J.; Zollhofer, M.; Stamminger, M.; Theobalt, C.; and Nie{\ss}ner, M.
  2016.
\newblock Face2{F}ace: Real-time face capture and reenactment of {RGB} videos.
\newblock In \emph{Proc. CVPR}, 2387--2395.

\bibitem[{Wang et~al.(2018{\natexlab{a}})Wang, Wang, Zhou, Ji, Gong, Zhou, Li,
  and Liu}]{wang2018cosface}
Wang, H.; Wang, Y.; Zhou, Z.; Ji, X.; Gong, D.; Zhou, J.; Li, Z.; and Liu, W.
  2018{\natexlab{a}}.
\newblock Cosface: Large margin cosine loss for deep face recognition.
\newblock In \emph{Proc. CVPR}, 5265--5274.

\bibitem[{Wang et~al.(2018{\natexlab{b}})Wang, Liu, Zhu, Tao, Kautz, and
  Catanzaro}]{Pix2PixHD}
Wang, T.-C.; Liu, M.-Y.; Zhu, J.-Y.; Tao, A.; Kautz, J.; and Catanzaro, B.
  2018{\natexlab{b}}.
\newblock High-resolution image synthesis and semantic manipulation with
  conditional gans.
\newblock In \emph{Proc. CVPR}, 8798--8807.

\bibitem[{Xu et~al.(2020)Xu, Yang, Chen, Wen, Deng, Jia, and Tong}]{xu2020deep}
Xu, S.; Yang, J.; Chen, D.; Wen, F.; Deng, Y.; Jia, Y.; and Tong, X. 2020.
\newblock Deep 3{D} Portrait from a Single Image.
\newblock In \emph{Proc. CVPR}, 7710--7720.

\bibitem[{Yu et~al.(2018)Yu, Wang, Peng, Gao, Yu, and Sang}]{yu2018bisenet}
Yu, C.; Wang, J.; Peng, C.; Gao, C.; Yu, G.; and Sang, N. 2018.
\newblock Bi{S}e{N}et: Bilateral segmentation network for real-time semantic
  segmentation.
\newblock In \emph{Proc. ECCV}, 325--341.

\bibitem[{Zeng et~al.(2020)Zeng, Pan, Wang, Zhang, and Liu}]{zeng2020realistic}
Zeng, X.; Pan, Y.; Wang, M.; Zhang, J.; and Liu, Y. 2020.
\newblock Realistic Face Reenactment via Self-Supervised Disentangling of
  Identity and Pose.
\newblock In \emph{Proc. AAAI}.

\bibitem[{Zhang et~al.(2019)Zhang, Huang, Li, Zhao, and Zhang}]{zhang2019multi}
Zhang, J.; Huang, Y.; Li, Y.; Zhao, W.; and Zhang, L. 2019.
\newblock Multi-attribute transfer via disentangled representation.
\newblock In \emph{Proc. AAAI}, volume~33, 9195--9202.

\bibitem[{Zhu et~al.(2020{\natexlab{a}})Zhu, Shen, Zhao, and
  Zhou}]{zhu2020domain}
Zhu, J.; Shen, Y.; Zhao, D.; and Zhou, B. 2020{\natexlab{a}}.
\newblock In-domain {GAN} inversion for real image editing.
\newblock In \emph{Proc. ECCV}.

\bibitem[{Zhu et~al.(2020{\natexlab{b}})Zhu, Abdal, Qin, and
  Wonka}]{zhu2020sean}
Zhu, P.; Abdal, R.; Qin, Y.; and Wonka, P. 2020{\natexlab{b}}.
\newblock {SEAN}: Image Synthesis with Semantic Region-Adaptive Normalization.
\newblock In \emph{Proc. CVPR}, 5104--5113.

\bibitem[{Zhu et~al.(2020{\natexlab{c}})Zhu, Xu, You, and
  Bai}]{zhu2020semantically}
Zhu, Z.; Xu, Z.; You, A.; and Bai, X. 2020{\natexlab{c}}.
\newblock Semantically Multi-modal Image Synthesis.
\newblock In \emph{Proc. CVPR}, 5467--5476.

\end{thebibliography}
}

\clearpage
 \begin{center}
	\huge \textbf{Appendix}
\end{center}

\subsection{1.Network architectures.}

\subsubsection{Style Encoder.} 
\begin{figure}[ht!]
	\centering
	\includegraphics[width=0.49\textwidth]{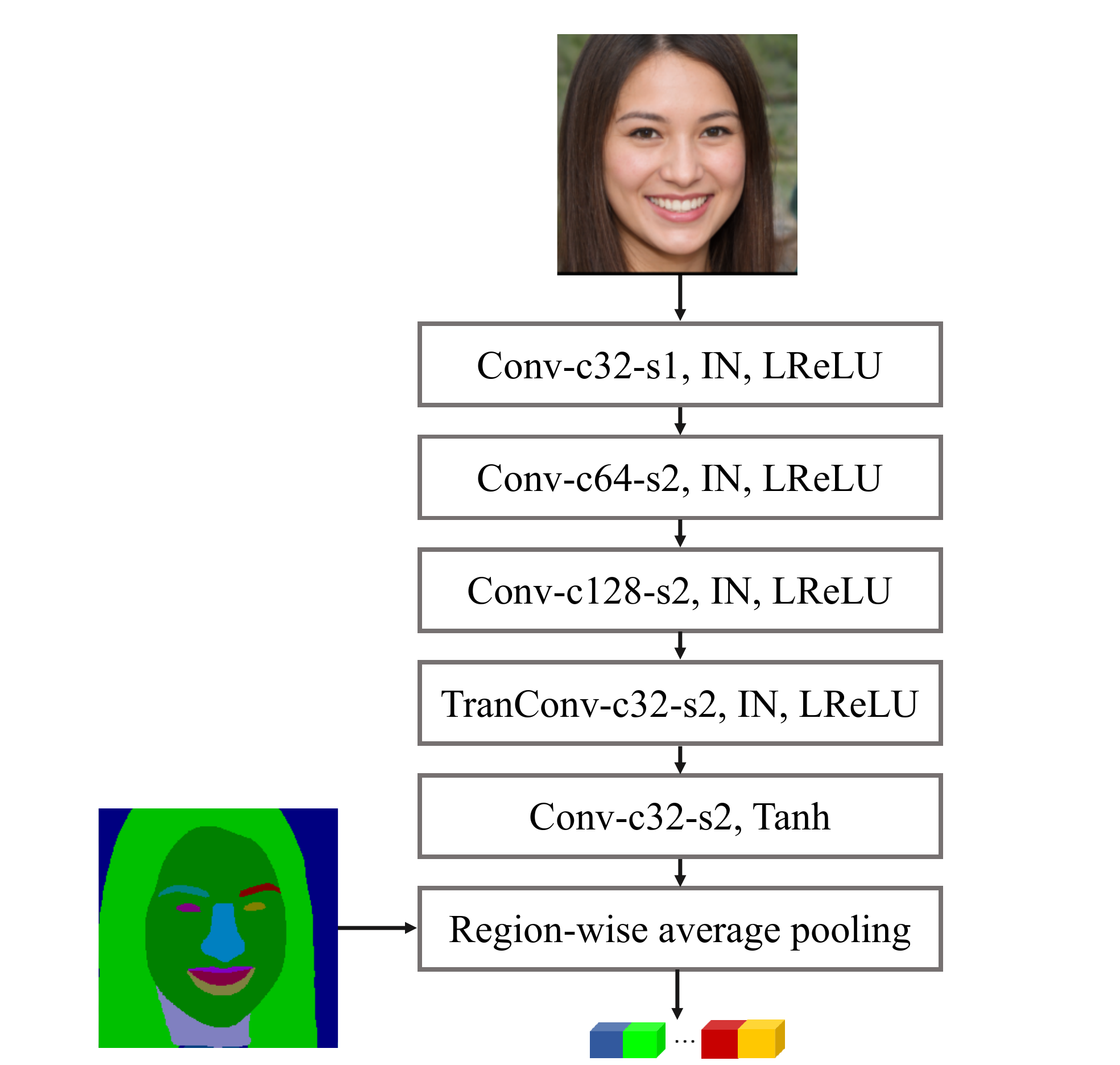}
	\caption{The architecture of style encoder. Note “Conv” and “TranConv” mean convolutional layer, transposed convolutional layer respectively. The numbers after “-c” and “-s” represent the channel number, stride the corresponding convolution. The kernel size of convolution is 3. “IN” represents instance normalization layer and “LReLU” means leaky ReLU layer.}
	\label{fig:style_encoder}
	\vspace{-2mm}
\end{figure}

We utilize the same architecture from SEAN~\cite{zhu2020sean} as our style encoder. The details can be seen in Fig.~\ref{fig:style_encoder}. To obtain the style matrix $S$, we employ region-wise average pooling at last layer output. Finally, we get $S \in \mathbb{R}^{B\times N \times D}$ to represent the style codes of each class. 

\subsubsection{Generator.} 
\begin{figure}[ht!]
	\centering
	\includegraphics[width=0.4\textwidth]{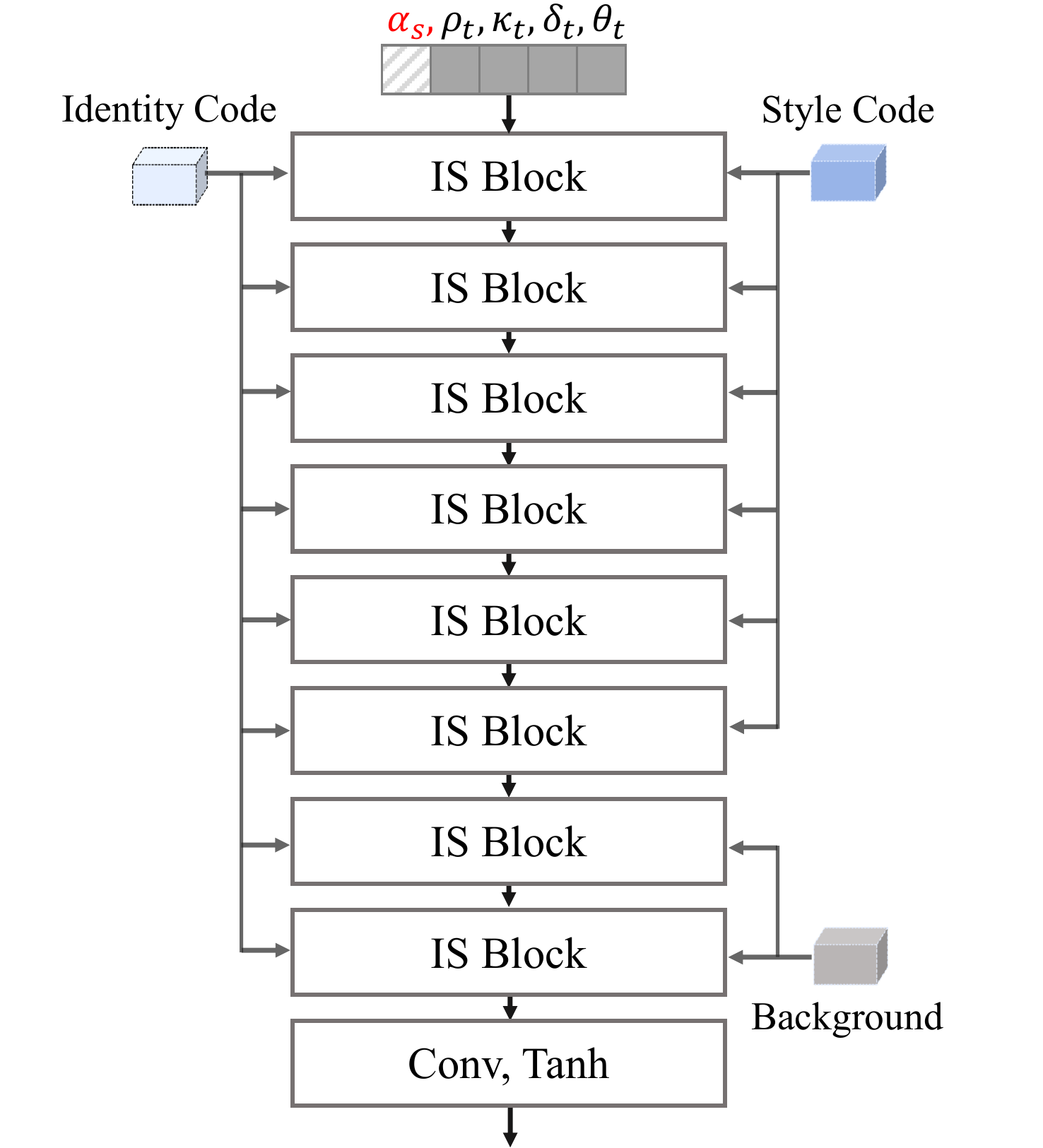}
	\caption{The architecture of our generator.}
	\label{fig:generator}
	\vspace{-4mm}
\end{figure}
The architecture of generator can be seen in Fig.~\ref{fig:generator}. The generator contains eight IS blocks which are mentioned above. We embed the background information into the last two blocks to keep the background unchanged with the target image.

\begin{figure}[ht!]
	\centering
	\includegraphics[width=0.40\textwidth]{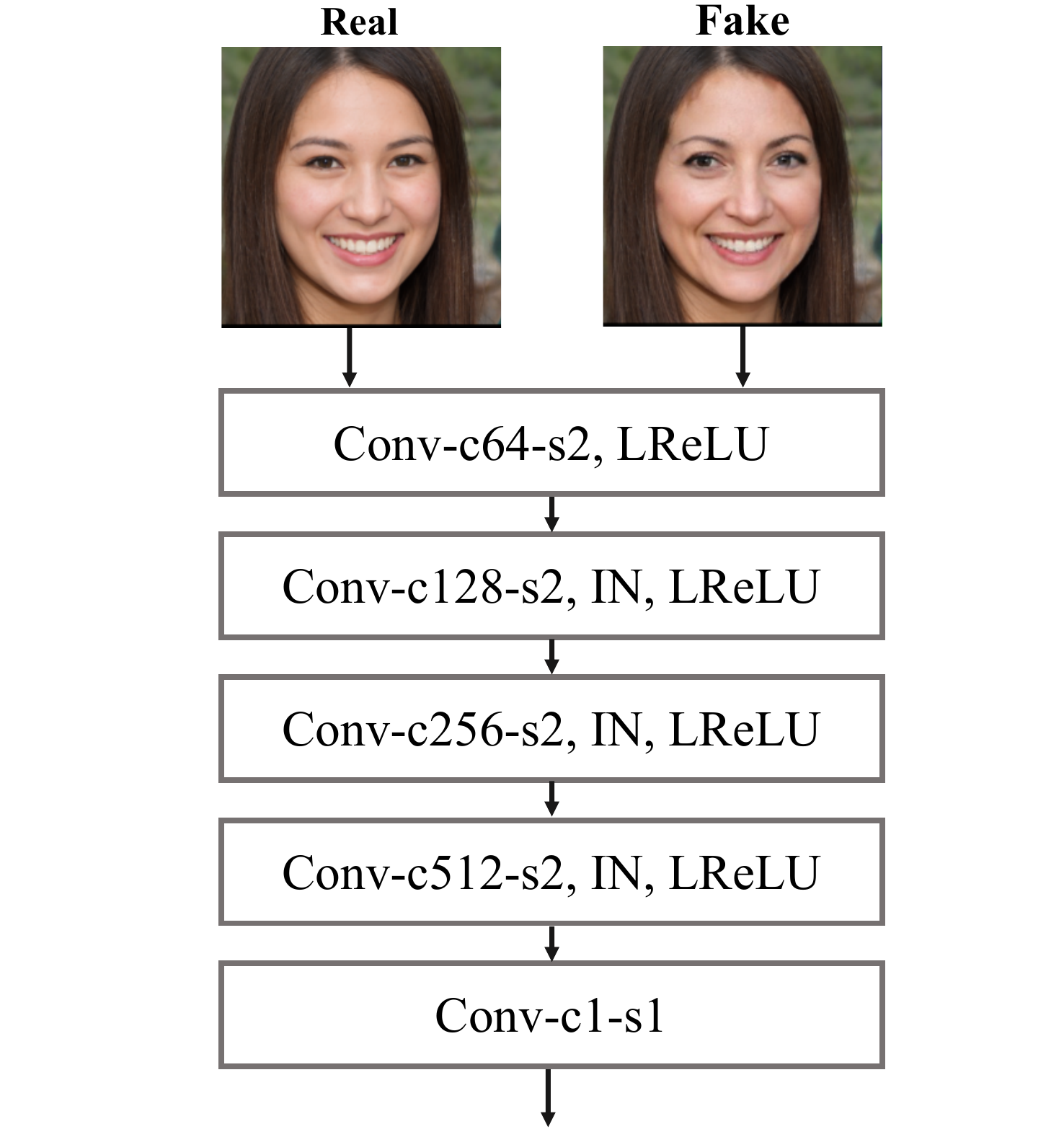}
	\caption{The architecture of our discriminator. Note that “Conv” denotes convolutional layer. The numbers after “-c” and “-s” represent the channel number and the stride of the corresponding convolutional layer. The kernel size of convolution is 4. “IN” represents instance normalization layer; “LReLU” detnotes leaky ReLU layer.}
	\label{fig:discriminator}
	\vspace{-4mm}
\end{figure}

\subsubsection{Discriminator.} 
We utilize the same discriminator architecture with Pix2PixHD~\cite{Pix2PixHD} and SPADE~\cite{SPADE}. There exist two multi-scale discriminator with instance normalization (IN) and Leaky ReLU (LReLU). The architecture of our discriminator is shown in Fig.~\ref{fig:discriminator}.

\subsection{2.Training details.}

We train the generator and the discriminator about 500K iterations. For the first 10K iteration, we drop out the landmark loss as it is unsteady when the generated face images are unnatural. Subsequently, when the generator can generate a face image, the landmark loss will be optimized steadily. The learning rate decay to 0 in the last 100K iterations. We apply spectral norm for the generator and discriminator following SPADE.
All experiments are conducted at one P40 GPU and the batch size is 16.

{\fontsize{9.0pt}{10.0pt} \selectfont

}

\end{document}